\documentclass[11pt,a4paper]{article}
\usepackage{times,latexsym}
\usepackage{url}
\usepackage[T1]{fontenc}
\usepackage{times}
\usepackage{latexsym}
\usepackage{enumitem}
\usepackage{multirow}
\usepackage{adjustbox}
\usepackage{subcaption}

\usepackage{booktabs}
\usepackage{array}
\usepackage{tablefootnote}
\usepackage{amsmath}
\usepackage{scalerel}
\usepackage{siunitx}
\usepackage{tabularx}
\usepackage[table, dvipsnames]{xcolor}
\usepackage{soul}
\usepackage{makecell}
\usepackage{algpseudocode}
\usepackage{inconsolata}
\usepackage{caption}

\usepackage{graphicx}
\usepackage{multirow}
\usepackage{adjustbox}

\usepackage{enumitem}
\usepackage{amsfonts}

\usepackage{siunitx}
\usepackage{dsfont}
\usepackage{amsmath}
\usepackage{amssymb}
\usepackage{bbm}
\usepackage{pifont}

\usepackage{tablefootnote}
\usepackage[utf8]{inputenc}
\usepackage{parskip}

\usepackage{color}
\usepackage{comment}
\usepackage{booktabs}

\usepackage{bbding}
\usepackage{algorithm}
\usepackage{algpseudocode}

\usepackage{courier}
\usepackage{url}

\newcommand{\ctext}[3][RGB]{%
  \begingroup
  \definecolor{hlcolor}{#1}{#2}\sethlcolor{hlcolor}%
  \hl{#3}%
  \endgroup
}

\usepackage[T1]{fontenc}

\usepackage[utf8]{inputenc}

\usepackage{microtype}

\usepackage{inconsolata}

\usepackage[acceptedWithA]{tacl2021v1}
\usepackage{xspace,mfirstuc,tabulary}

\newif\iftaclinstructions
\taclinstructionsfalse 
\iftaclinstructions
\renewcommand{\confidential}{}
\renewcommand{\anonsubtext}{(No author info supplied here, for consistency with
TACL-submission anonymization requirements)}
\newcommand{\instr}
\fi

\iftaclpubformat 

\else

\fi

\title{JustiLM: Few-shot Justification Generation for Explainable Fact-Checking of Real-world Claims}

\author{Fengzhu Zeng 
  \\
  Singapore Management University
  \\
  80 Stamford Rd, Singapore 178902
  \\
  \texttt{fzzeng.2020@phdcs.smu.edu.sg}
  \And
  Wei Gao 
  \\
  Singapore Management University
  \\
  80 Stamford Rd, Singapore 178902
  \\
  \texttt{weigao@smu.edu.sg}
}

\date{}

\begin{document}
\maketitle
\begin{abstract}
Justification is an explanation that supports the veracity assigned to a claim in fact-checking. However, the task of justification generation is previously oversimplified as summarization of fact-check article authored by fact-checkers. Therefore, we propose a realistic approach to generate justification based on retrieved evidence. We present a new benchmark dataset  called ExClaim for \underline{Ex}plainable fact-checking of real-world \underline{Claim}s, and introduce JustiLM, a novel few-shot \underline{Justi}fication generation based on retrieval-augmented \underline{L}anguage \underline{M}odel by using fact-check articles as auxiliary resource during training only. Experiments show that JustiLM achieves promising performance in justification generation compared to strong baselines, and can also enhance veracity classification with a straightforward extension.~\footnote{Code and dataset are released at \url{https://github.com/znhy1024/JustiLM}}

\end{abstract}

\section{Introduction} \label{intro}

\begin{table*}[t!]
    \scriptsize
    \begin{tabularx}{\linewidth}{m{1.5cm}|X}
        \toprule
        \textbf{Claim} & Biden: Gun manufacturers are “the only industry in the country” that have immunity from lawsuits. \\
        \midrule
        \textbf{Evidence Documents\newline (References)} & \ctext[RGB]{186,248,255}{Doc1:} \textit{No, you can't sue Pfizer or another manufacturer if you get a COVID-19 vaccine injury, but you can file for compensation.}  The Pfizer-BioNTech COVID-19 vaccine received full approval from the Food and Drug ... \newline
        \ctext[RGB]{218,224,236}{Doc2:} \textit{Remarks by President Biden on Gun Violence Prevention.} THE PRESIDENT: Thank you, Kamala — Madam Vice President. Thank you very much.  You know, we’re joined ... \newline 
        \ctext[RGB]{238,206,206}{Doc3:} \textit{Clinton: Gun industry is `wholly protected' from all lawsuits.} At the first Democratic debate of the 2016 presidential race, former Secretary of State Hillary Clinton criticized opponent ... \newline 
        \ctext[RGB]{220,233,213}{Doc4:} \textit{Protection of Lawful Commerce in Arms Act.} The Protection of Lawful Commerce in Arms Act (PLCAA) is a United States law which protects firearms manufacturers and dealers from being held liable ... \newline
        ... \\
        \midrule
        \textbf{Fact-check Article} & ... \ctext[RGB]{218,224,236}{This isn't the first time Biden has made this claim. He's made it repeatedly, including April 2021 remarks about gun violence} ... \textbf{But the claim is inaccurate. The gun industry is susceptible to some lawsuits, and there are federal laws restricting liability for a number of other types of businesses.} ... \ctext[RGB]{220,233,213}{The law says gun dealers and manufacturers cannot be sued when their products are misused. But the law lists several situations that are not protected from liability.} ... \ctext[RGB]{220,233,213}{Other industries have exemptions in liability.} ... \ctext[RGB]{186,248,255}{until 2024, pharmaceutical companies that make the COVID-19 vaccines will have liability immunity} ... \ctext[RGB]{238,206,206}{\textbf{There's also some liability protection in the medical devices and airline industries.}} ... \\
        \midrule
        \textbf{Justification} & Biden said that gun manufacturers represent the only industry in America that is exempt from being sued. This isn't accurate. The gun industry is not entirely exempt from being sued and is susceptible to some lawsuits. Further, there are federal laws that restrict liability for a variety of other business sectors. We rate it False. \\
        \midrule
        \textbf{Veracity} & FALSE \\
        \midrule
        \textbf{Procedure} & \raisebox{-.9\height}{\includegraphics[width=\linewidth,keepaspectratio]{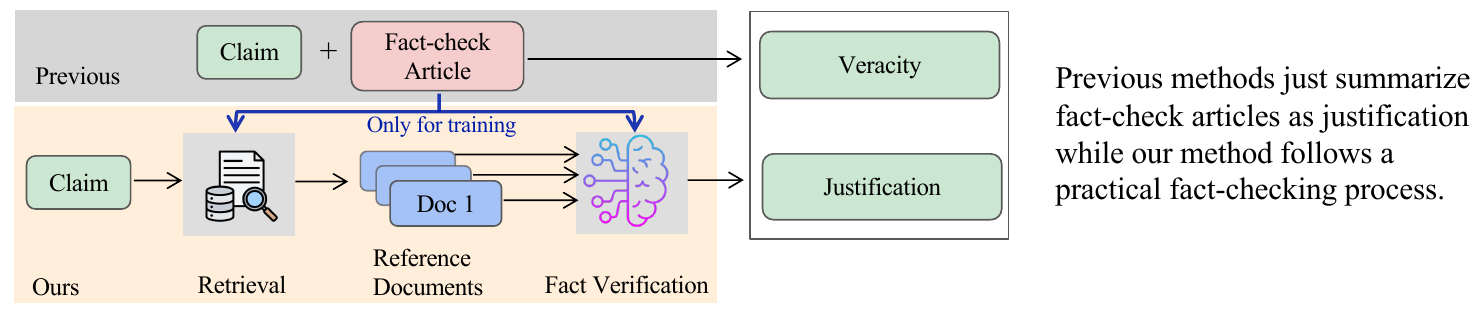}} \\
        \bottomrule
    \end{tabularx}
    \caption{An example claim along with the evidence documents, justification and veracity. The \textit{title} of each evidence document is italicized. The sentences in the fact-check article referring to evidence documents are marked in the same color as the corresponding documents, and the sentences that directly entail the justification are in \textbf{bold}.}
    \label{tab:intro example}
\end{table*}

Automated fact-checking typically encompasses several stages: identify check-worthy claims, retrieve relevant evidence, determine the claim's veracity using the retrieved evidence, and generate justification for the verdict on the veracity~\citep{guo-etal-2022-survey}. Despite a wealth of research focusing on the initial three stages, justification generation remains under-explored in the past. Justifications present essential evidence and rationales used to arrive at a claim's veracity judgement, serving to convince readers and enhance the credibility of fact-checking systems. This explanatory process is of paramount importance in gaining user's trust in automated fact-checking~\citep{kotonya-toni-2020-explainable,atanasova-etal-2020-generating-fact}. 

Several methods have attempted to generate justification of verdict by summarizing fact-check articles that were previously authored by human fact-checkers~\citep{kotonya-toni-2020-explainable-automated,atanasova-etal-2020-generating-fact,Benchmarking}. 
Since a fact-check article per se is manually written to justify the verdict of given claim with detailed presentation and reasoning over digested evidence, referring to reference documents collected from multiple sources, directly generating a summary from such report as justification sidesteps the realistic challenges of evidence gathering and evidence-based reasoning for veracity assessment we essentially face in fact-checking task. More importantly, these existing methods are impractical because fact-check articles are not available for new claims that are yet to check~\cite{guo-etal-2022-survey}. Table~\ref{tab:intro example} shows an example illustrating different types of information involved in the fact-checking practice and their relationship. To justify the veracity for a claim, the source of information that can be used practically ought to be the retrieved reference documents containing evidence rather than its fact-check article, which, as an outcome, has not been written during the checking process.

In this paper, we propose a more realistic approach for the task of justification generation based on a language model approach, which complies with the process of journalistic fact-checking by well-known fact-check organizations such as PolitiFact\footnote{\url{https://www.politifact.com/}}. 
Our goal is to produce high-quality justifications, drawing upon evidence gathered from diverse sources.
To this end, we construct a benchmark dataset for \underline{Ex}plainable fact-checking of real-world \underline{Claim}s, named ExClaim, derived from a public dataset WatClaimCheck~\cite{khan-etal-2022-watclaimcheck} containing newsworthy claims along with their fact-check articles and reference documents. ExClaim provides a large searchable corpus by mixing the reference documents from all claims in WatClaimCheck. Additionally, it curates the verdict justifications sourced from fact-check articles, typically located in a conclusive paragraph marked by cue phrases like ``Our ruling'' or ``Our rating'' for each claim. 
Furthermore, we develop a \underline{Justi}fication \underline{L}anguage \underline{M}odel called JustiLM for generating the rationales behind veracity judgement within the context of few-shot learning. Presumably, few-shot fine-tuning can mitigate the training resource requirements and its dependence on high-end hardware, often financially prohibitive, and also enables the model to achieve comparable effectiveness to state-of-the-art fully-trained models. JustiLM utilizes fact-check articles as auxiliary information in its training only via fine-tuning a pre-trained Retrieval-Augmented Generation (RAG) model on our curated justification dataset. Meanwhile, leveraging fact-check articles for training enhances the model's proficiency in generating rationales based on evidence and articulating them in its generated content. 
Our contributions are threefold:
\begin{itemize}[leftmargin=*]
    \item We propose JustiLM, the first realistic justification generation method based on a retrieval-augmented language model that is trained end-to-end for explainable fact checking of real-world claims, leveraging fact-check articles as auxiliary information for model training only. 
    
    \item We construct ExClaim, a new benchmark  derived from the WatClaimCheck dataset~\cite{khan-etal-2022-watclaimcheck} for the explainable fact-checking, which contains 6,951 real-world claims and their corresponding veracity labels and human-written justifications, together with a large searchable corpus of 957,949 chunk-level documents for fine-grained evidence retrieval.
    
    \item JustiLM outperforms In-Context Learning (ICL) enabled language models, including Flan-T5, Llama2, and the state-of-the-art few-shot RAG model Atlas. JustiLM also shows promising performance compared to the latest GPT-4 model. A straightforward extension of JustiLM for joint veracity prediction and justification generation improves the veracity prediction task with large margins.
\end{itemize}

\section{Related Work}
\subsection{Explanations for fact-checking} \label{relate work: exp fc}
Explanations for fact-checking claims have gained significant prominence in recent times, particularly due to the prevalent use of black-box models in automated fact-checking systems~\citep{atanasova-etal-2020-generating-fact,guo-etal-2022-survey}. Several methods have emerged to address this issue utilizing various techniques to provide human readable explanations. One stream of research leverage attention weights to highlight salient parts in the retrieved evidence as explanations~\citep{popat-etal-2018-declare,ma-etal-2019-sentence,XFake,dEFEND,lu-li-2020-gcan}. Another stream of works is to adopt logic-based rules, such as knowledge graphs and natural logic relations designed by human experts~\citep{AhmadiLPS19,ExFaKT,FACE-KEG,ProoFVer}, where explanations are obtained by tracing the rules path to reach the veracity of the claim. However, these explanations are not presented in natural language, rendering them less accessible to general users. Furthermore, these rule-based systems encounter challenges when dealing with real-world claims that may not conform to predefined rules. In contrast, our work places a strong emphasis on generating textual justifications that are readily understandable for users, avoiding manual rule definitions.

A few studies have attempted to automatically generate textual justifications by summarizing fact-check articles~\citep{kotonya-toni-2020-explainable-automated,atanasova-etal-2020-generating-fact, Benchmarking}. \citet{atanasova-etal-2020-generating-fact} employs DistilBERT~\cite{sanh2019distilbert} to extract sentences from fact-check articles to form justifications. \citet{kotonya-toni-2020-explainable-automated} proposes a two-step process, initially utilizing a Sentence-BERT~\cite{reimers-gurevych-2019-sentence} to extract sentences from fact-check articles and subsequently using the BERTSUM model~\cite{liu-lapata-2019-text} for abstractive justification generation based on the extracted sentences. \citet{Benchmarking} explores several existing extractive summarization~\citep{LexRank,reimers-gurevych-2019-sentence} and abstractive summarization~\citep{raffel2020exploring,PEGASUS,dBart} approaches for summarizing fact-check articles.
These summarization methods come with inherent limitations practically, including complete reliance on fact-check articles (i.e., detailed human justification) as input, which is hardly available at the time of deployment, and complete omission of automatic evidence search and evidence-based reasoning. Different from these approaches, our method only assumes the availability of fact-check articles during model training and the key evidence exists within a large corpus which is searchable. Therefore, our approach generates justifications by harnessing the information from retrieved reference documents during inference, which is a more realistic solution for real-world scenarios. Similarly, \citet{khan-etal-2022-watclaimcheck} infers claim veracity based on retrieved textual references, while \citet{end2end-mm} retrieves evidence for multi-modal fact-checking and generates explanations for predicted veracity labels using the BART model~\cite{lewis-etal-2020-bart}, both of which are stage-wise and full-dataset trained. In contrast, we base our approach on the latest RAG framework that is trained end-to-end and generates justifications by using fact-check articles to distill supervisory signals for training. 

\subsection{Few-shot fact-checking}
The need of few-shot learning is exacerbated by the continuous increase of computational and storage requirements for language model training. However, the specific application of few-shot learning techniques in the context of fact-checking has been relatively underexplored.
Existing methods for few-shot fact-checking only focus on the so-called fact verification task~\citep{lee-etal-2021-towards,zeng2022aggregating,zeng2023prompt,yue2023metaadapt,pan2023factchecking,zhang-gao} by feeding a few instances together with gold evidence into the model to predict the veracity of a claim. Different from these methods, our work primarily centers on generating justifications to substantiate the veracity of a claim based on the \emph{retrieved} evidence. Importantly, we do not assume the availability of annotated evidence. Instead, we necessitate the system to retrieve pertinent evidence, conforming to a more realistic and challenging scenario. 

\subsection{Retrieval-augmented language models}

Equipping language models (LM) with external memory has shown to enhance their performance in knowledge intensive NLP tasks~\citep{chen-etal-2017-reading,thorne-etal-2018-fever,REALM,rag,SachanRHDY21,izacard-grave-2021-leveraging,BorgeaudMHCRM0L22,atlas}. Typically, a retriever is used to retrieve relevant documents from a large corpus, which enriches the input of a language model and contributes to the final output. However, due to the high cost of acquiring query-document annotations and training retrievers, many implementations rely on off-the-shelf retrievers, such as TF-IDF and BM25~\citep{tfidf,Okapi-BM25}, which use term-matching techniques. In this setup, only the parameters of LMs are fine-tuned.

Recent research has demonstrated the advantages of jointly training the retriever and the LM in an end-to-end manner, which leverages the supervision signals from the LM to train the retriever~\citep{REALM,rag,SachanRHDY21,izacard-grave-2021-leveraging,atlas}.
Moreover, considering the remarkable performance of large language models (LLMs) in various few-shot NLP tasks, some studies suggest enhancing LLMs with the retrievers or web search engines~\citep{mallen-etal-2023-trust,SiGYWWBW23,Generate-rather-than-Retrieve,replug,zhang-gao}. For example, REPLUG~\cite{replug} optimizes the retriever by minimizing the KL divergence between the retrieval likelihood and the black-box LLM likelihood over retrieved documents. However, there exists inherent limitations in the interaction between retriever and black-box LLMs, such as their restricted ability to provide or access specific information.
We refer readers to a comprehensive survey of retrieval-augmented LMs~\cite{Augmented-LM-survey}. 

\section{Task Formulation} \label{Task Formulation}
Let $\mathbf{C}= \{(\mathbf{x},\mathbf{z},\mathbf{y})\}$ be a fact-checking dataset of real-world news claims associated with a textual knowledge corpus $\mathcal{D}$.
Each instance is composed of a claim $\mathbf{x}$ and its corresponding ground-truth justification $\mathbf{y}$ and fact-check article $\mathbf{z}$. 
$\mathbf{C}$ is divided as a training set and a test set, and only instances in the training set are associated with fact-check articles if available.

Given a claim $\mathbf{x}$ and the corpus $\mathcal{D}$, the goal of justification generation is to produce a sequence of tokens, denoted as $\hat{\mathbf{y}}$, that serves as an explanation for the veracity rendered on the claim using the evidence retrieved from the corpus. In the few-shot setting, we randomly select $K$ instances from the training set, following the similar setup employed in previous studies for fact verification~\cite{lee-etal-2021-towards,liu2022few,zeng2023prompt}, and we do not assume the availability of development set as this aligns to a more realistic scenario with limited data resources.

\begin{table}[t]
\centering
\small
\normalsize
\begin{adjustbox}{width={\linewidth},totalheight={!},keepaspectratio}%
\begin{tabular}{llcc}
\toprule[1.0pt]
 & Split &\# Instance   & Avg. \# Tokens.  \\ 
\midrule[0.5pt]
\multirow{2}{*}{Claim} 
&Train  & $5,964$       &  $25$  \\
&Test  & $987$       & $25$  \\
\midrule[0.5pt]
\multirowcell{2}[0pt][l]{Fact-check \\Article}  
&Train  &  $5,964$      &  $1,102$  \\
&$\text{Test}^{\dag}$  &  $987^{\dag}$     & $1,091^{\dag}$  \\
\midrule[0.5pt]
\multirowcell{2}[0pt][l]{Reference \\Documents} 
&Train  &  $40,089$    & $2,656$  \\
&Test  &  $6,647$     & $2,404$  \\
\midrule[0.5pt]
\multirow{2}{*}{Justification}  
&Train  &   $5,964 $    & $129$   \\
&Test  & $987$     & $131$   \\
\bottomrule[1.0pt]
\end{tabular}
\end{adjustbox}
\caption{Statistics of the ExClaim dataset. \dag: Note that fact-check articles in the test set are not used in our method, but exclusively utilized by baselines that rely on fact-check articles.
}
\label{tab:dataset statistics}
\end{table}

\section{ExClaim Dataset}~\label{sec:dataset}
The existing fact-checking datasets based on real-world claims have limitations for justification generation. This is because the provided evidence sources might not cover the evidence documents that fact-checkers actually rely on when writing justifications. For example, some datasets~\citep{vlachos-riedel-2014-fact,wang-2017-liar,alhindi-etal-2018-evidence} only provide metadata like speaker, party and date without a sizeable knowledge corpus for finding specific evidence. Some studies ~\citep{CreditAssess,baly-etal-2018-integrating,augenstein-etal-2019-multifc,gupta-srikumar-2021-x,yang-etal-2022-coarse,hu-etal-2022-chef} utilize web search to gather evidence documents, which result in retrieved information from non-authoritative sources or lead to the leak of ground truth by inadvertently including articles verifying the same claims by other organizations or sharing the fact-check information~\cite{khan-etal-2022-watclaimcheck}.
More notably, certain studies~\citep{hanselowski-etal-2019-richly,kotonya-toni-2020-explainable,atanasova-etal-2020-generating-fact,Multi-Hop,Benchmarking} regard fact-check articles as primary source of evidence, a practice that may not align with the realistic fact-checking procedures.

We use the WatClaimCheck~\cite{khan-etal-2022-watclaimcheck} dataset that provides the real-world claims along with the text of reference documents cited by fact-check articles. However, WatClaimCheck is constructed for veracity classification and does not provide ground-truth justifications. 
For our task, we construct ExClaim based on WatClaimCheck, for which we additionally extract justifications from fact-check articles based on the cue phrases such as "Our ruling" or "Our rating" in the reports following previous works~\citep{alhindi-etal-2018-evidence,augenstein-etal-2019-multifc,kotonya-toni-2020-explainable} and remove the instances that do not have such justification content. After extracting the justifications, we also remove them from fact-check articles.

Table \ref{tab:dataset statistics} presents summary statistics of ExClaim dataset with total 6,951 real-world claims and justifications (i.e., 5,964 for training and 987 for testing). The data pose some challenges: 1) A single reference document is generally much longer than fact-check article, easily exceeding the context window of most text generation models (e.g., 512 tokens of T5~\cite{raffel2020exploring} or 1,024 tokens of BART~\cite{lewis-etal-2020-bart}). In particular, each claim may correspond to multiple reference documents from different sources, leading to excessively long text for evidence. 2) There is lack of passage-/sentence-level annotation in reference documents and fact-check articles. Since fact-checkers generally refer to only several pieces of texts in reference documents when writing justifications, most information in a reference document tend to be irrelevant for generating the justifications. To address these issues, we split each document into disjoint 100-word chunks following previous works~\citep{lee-etal-2019-latent,karpukhin-etal-2020-dense,rag,atlas}, resulting in a large textual knowledge corpus $\mathcal{D}$ comprising a total of 957,949 chunk-level documents that systems can search fine-grained evidence text from. In the rest of the paper, we refer to these short text chunks as ``reference documents'' or simply ``documents''. 

\section{Methodology}

\begin{figure*}[t!]
  \centering
  \includegraphics[width=0.9\textwidth]{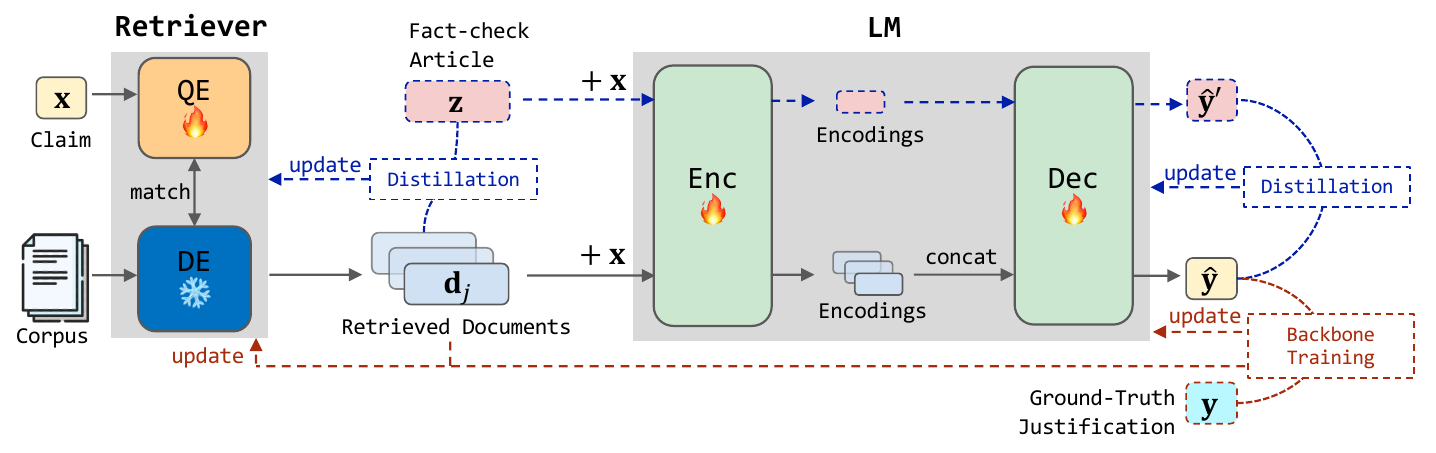}
    \caption{The architecture of JustiLM. Grey solid arrows present the inference process \textbf{without} fact-check article $\textbf{z}$. Red dash arrows present the training process of backbone model, where the ground-truth justification provide supervisory signals to train both retriever and LM. Blue dash arrows present the training process with the distillation of $\textbf{z}$ as supervisory signals. The document encoder is fixed during training, while other modules are trainable. QE: Query Encoder; DE: Document Encoder; Enc: Encoder; Dec: Decoder.}
  \label{fig:model}
\end{figure*}

We base our approach on the retrieval-augmented generation (RAG) framework~\citep{rag,SachanRHDY21,izacard-grave-2021-leveraging,atlas}, which contains a retriever for fine-grained evidence retrieval and a LM for textual justification generation. As shown in Figure \ref{fig:model}, the retriever takes the claim text as input and retrieves the top-$N$ chunk-level documents from the textual knowledge corpus, and the LM conditions on these documents together with the claim to generate justification. 
The retriever and LM can be jointly trained within a single RAG framework, which makes it possible to utilize fact-check articles as auxiliary resource to provide supervisory signals during training, targeting to enhance the quality of generated justification. We employ Atlas~\cite{atlas} as our backbone model considering two main reasons: 1) its strong few-shot learning ability in knowledge intensive tasks when its retriever and LM are jointly trained; 2) its flexibility for incorporating fact-check articles in the training process. 

\subsection{Retriever} \label{retr}
Given a claim $\mathbf{x}$, the retriever should return the documents that help LM generate better justification. To enable the training of retriever, Atlas utilizes a dense retriever named Contriever~\cite{contriever}, which is pre-trained using the MoCo contrastive loss~\cite{moco}. Contriever is a dual-encoder architecture that the pre-trained query encoder $\textbf{E}_c$ and document encoder $\textbf{E}_d$ encode the claim $\mathbf{x}$ and each document $\mathbf{d}_{j} \in \mathcal{D}$, respectively. The embeddings of documents can be pre-computed to build a collection of index using FAISS~\cite{faiss} for fast retrieval. Documents are ranked by the similarity score $s(\mathbf{x},\mathbf{d}_{j})=\textbf{E}_c(\mathbf{x})^\top\textbf{E}_d(\mathbf{d}_{j})$ that is calculated by taking the dot product of the embeddings of the claim $\mathbf{x}$ and document $\mathbf{d}_{j}$. 

To mitigate the burden of re-computing embeddings for all documents when training the retriever, Atlas~\cite{atlas} only updates the parameters corresponding to the query encoder while freezing the documents encoder, which still shows promising results in the few-shot setting. Therefore, we employ the document encoder for encoding reference documents and the query encoder for encoding other inputs. Since there is no direct supervision available to train the retriever, Atlas proposes a Perplexity Distillation loss to leverage the supervisory signals from the LM. The intuition behind is that documents contributing to the LM that help generate lower-perplexity outputs should be ranked higher~\cite{atlas}. 

\subsection{Language Model}  \label{lm}

The language model conditions on the top-$N$ retrieved documents $D_N=\{\mathbf{d}_j\}_{j=1}^{N}$ by the retriever, together with the claim $\mathbf{x}$, to generate the justification. To aggregate evidence efficiently and effectively from multiple documents in LM, Atlas employs a T5 encoder-decoder model~\cite{raffel2020exploring} with the Fusion-in-Decoder (FiD)~\cite{izacard-grave-2021-leveraging} modification. 
Each retrieved document $\mathbf{d}_{j}$ is encoded independently by the encoder, with the claim $\mathbf{x}$ prepended to it. 
All outputs of the encoder are then concatenated. The decoder takes as input this concatenation and performs cross-attention to fuse the evidence and generate outputs. The training objective is the standard language modeling loss that encourages the LM to assign higher probability to the target sequence $\mathbf{y}$ given the claim $\mathbf{x}$ and top-$N$ retrieved documents.

\subsection{Distillation Techniques}
Although directly summarizing fact-check articles $\mathbf{z}$ can generate justifications with reasonable quality in previous works~\citep{kotonya-toni-2020-explainable,atanasova-etal-2020-generating-fact}, $\mathbf{z}$ is by no means available during inference for new claims in real-world deployment, as we discussed in $\S$\ref{intro}, making the previous methods impractical. We propose a realistic approach to address this limitation: distilling information from $\mathbf{z}$ as auxiliary supervisory signals for training phase only. We introduce two types of techniques based on the granularity of distillation from fact-check articles. The first is article-level distillation, which utilizes aggregated information from the entire $\mathbf{z}$. The second is chunk-level distillation, where we split each article $\mathbf{z}$ as multiple disjoint 100-word chunks $\mathbf{z}=\{\mathbf{z}_{i}\}_{i=1}^{M}$, where $M=\lceil \frac{|\mathbf{z}|}{100}\rceil$. Chunk-level distillation utilizes individual information of each chunk $\mathbf{z}_{i}$. Both types of distillation techniques can be applied to train the retriever and LM.

\subsubsection{Article-level Distillation} \label{global}
Article-level distillation is performed at the entirety of a fact-check article, aiming at utilizing the global-level alignment between fact-check article $\mathbf{z}$ and retrieved documents $D_N$ as supervisory signals for model training. The basic idea is that the more similar $D_N$ and $\mathbf{z}$ are, the easier it is for LM to generate justification based on $D_N$ closely approximating that generated based on $\mathbf{z}$. This alignment serves two main purposes. Firstly, the similarity between $D_N$ and $\mathbf{z}$ can act as a supervisory signal, guiding the retriever to prioritize the ranking of documents in $D_N$ to resemble $\mathbf{z}$. Secondly, the justification generated by the LM based on $\mathbf{z}$ can be used as a supervision signal to encourage the LM using $D_N$ to generate justification as similar as those generated based on $\mathbf{z}$. Next, we will discuss two training losses that serve both purposes.

\textbf{Retrieval loss.}
The technique for training retriever is based on the similarity between the \textit{entire} fact-check article $\mathbf{z}$ and retrieved documents $D_N$. However, the length of $\mathbf{z}$ is commonly larger than the maximum input length (i.e., 512 tokens) of query encoder. Therefore, we use the trainable query encoder $\textbf{E}_c$ to represent $\mathbf{z}$ by aggregating the embeddings of all its chunks and obtain $\bar{\textbf{E}}_c(\mathbf{z})=\frac{1}{M}\sum_{i=1}^{M} {\textbf{E}}_c(\mathbf{z}_{i})$. The training objective is to minimize the mean-squared-error (MSE) loss between the embeddings of $\mathbf{z}$ and  $\mathbf{d}_i$:
\begin{equation}  
\mathcal{L}^{\mathrm{ret}}_{g}=\frac{1}{N | \bar{\textbf{E}}_c(\mathbf{z})|}\sum_{j=1}^{N} ||\bar{\textbf{E}}_c(\mathbf{z})-\textbf{E}_d(\mathbf{d}_{j})||_2^2.
\end{equation}
\textbf{Generation loss.} The technique for training the LM generation is based on the distance between the generated justification using retrieved documents $D_N$ and that directly using the fact-check article $\mathbf{z}$. During training, the generation $\hat{\textbf{y}}$ of the LM using $\mathbf{z}$ as input is regarded as supervision signal to guide model's learning.  Let $p_{\scaleto{\mathrm{L}}{4pt}}(\mathbf{y} \mid \mathbf{x},D_N)=\prod_{k=1}^{|\mathbf{y}|} p_{\scaleto{\mathrm{L}}{4pt}}(t_k \mid \mathbf{x},D_N,t_{<k})$ be the LM probability of generating the ground-truth justification $\mathbf{y}$ conditioned on $\mathbf{x}$ and $D_N$, where $p_{\scaleto{\mathrm{L}}{4pt}}(t_k \mid \mathbf{x},D_N,t_{<k})$ is the probability of each token $t_k$ assigned by the LM and $t_{<k}$ denotes the tokens generated prior to $t_k$.  Similarly, the LM probability of generating $\mathbf{y}$ conditioned on $\mathbf{z}$ is $p_{\scaleto{\mathrm{L}}{4pt}}(\mathbf{y} \mid \mathbf{x},\mathbf{z})$. The training objective is to minimize the MSE loss between these two distributions:
\begin{align}    \mathcal{L}^{\mathrm{lm}}_{g}&=\frac{1}{|\mathbf{y}||\mathcal{V}|} \sum_{k=1}^{|\mathbf{y}|}\sum_{i=1}^{|\mathcal{V}|}(p_{\scaleto{\mathrm{L}}{4pt}}(t_i \mid \mathbf{x},D_N,t_{<k})\\ \nonumber
 &-p_{\scaleto{\mathrm{L}}{4pt}}(t_i \mid \mathbf{x},\mathbf{z},t_{<k}))^2,
\end{align}
where $\mathcal{V}$ is the vocabulary of the LM.

\subsubsection{Chunk-level Distillation} \label{chunk}
Chunk-level distillation is performed at the granularity of each chunk of fact-check article, leveraging the alignment between chunks $\{\mathbf{z}_{i}\}_{i=1}^{M}$ and documents $\{\mathbf{d}_j\}_{j=1}^{N}$ to provide supervisory signals for model training. The  intuition is that different chunks of the fact-check article could be derived from rearranging or modifying specific text spans sourced from reference documents. Further, the chunks $\{\mathbf{z}_{i}\}_{i=1}^{M}$ may correspond to certain parts of the ground-truth justification $\mathbf{y}$. Thus, $\{\mathbf{z}_{i}\}_{i=1}^{M}$ can be seen as the "connections" between $D_N$ and $\mathbf{y}$. Aligning $\{\mathbf{d}_j\}_{j=1}^{N}$ and $\{\mathbf{z}_{i}\}_{i=1}^{M}$ intuitively aids the model in learning the mapping from $D_N$ to $\mathbf{y}$, hence improving its performance. However, there is no chunk-level annotation available, which poses an important challenge for training. We design two training techniques to address it for chunk-level distillation in both retriever and LM.

\textbf{Retrieval loss.}  
The technique for training the retriever is based on the relation between similarity score and the LM perplexity, which is inspired by~\citet{atlas} and~\citet{replug}. Intuitively, the more similar the text chunk $\mathbf{z}_{i}$ is to the document $\mathbf{d}_{j}$, the lower LM perplexity of generating $\mathbf{z}_{i}$ conditioned on $\mathbf{d}_{j}$:
\begin{equation*}  
s(\mathbf{z}_{i},\mathbf{d}_{j}) \propto p_{\scaleto{\mathrm{L}}{4pt}}(\mathbf{z}_{i} \mid \mathbf{x},\mathbf{d}_j),
\end{equation*}
where $s(\mathbf{z}_{i},\mathbf{d}_{j})=\textbf{E}_c(\mathbf{z}_{i})^\top\textbf{E}_d(\mathbf{d}_{j})$. We train the retriever to learn the alignment between $\mathbf{d}_{j}$ and its most similar chunk $\mathbf{z}_{j^*}$, where $j^*=\arg\max_{i\in [1,M]}s(\mathbf{z}_{i},\mathbf{d}_{j})$. It involves minimizing the the KL-divergence between the similarity score $s(\mathbf{z}_{j^*},\mathbf{d}_{j})$ and the corresponding LM probability of $\mathbf{z}_{j^*}$ conditioned on $\mathbf{d}_j$ and $\mathbf{x}$. Specifically, let the documents distribution over $D_N$ be $p_{\scaleto{\mathrm{R}}{4pt}}(\mathbf{d}_j \mid \mathbf{z}_{i})= \frac{\text{exp}(s(\mathbf{z}_{i},\mathbf{d}_{j}))}{\sum_{k=1}^{N}\text{exp}(s(\mathbf{z}_{i},\mathbf{d}_{k}))}$, and the documents posterior distribution according to the LM be $q_{\scaleto{\mathrm{L}}{4pt}}(\mathbf{z}_{j^*} \mid \mathbf{x},\mathbf{d}_j)=\frac{\text{exp}(\log p_{\scaleto{\mathrm{L}}{4pt}}(\mathbf{z}_{j^*} \mid \mathbf{x},\mathbf{d}_j))}{\sum_{k=1}^{N}\text{exp}(\log p_{\scaleto{\mathrm{L}}{4pt}}(\mathbf{z}_{j^*} \mid \mathbf{x},\mathbf{d}_k))}$. Finally, the loss function for optimizing the retriever is given as:
\begin{equation}   
\mathcal{L}^{\mathrm{ret}}_{c}=\sum_{j=1}^{N} q_{\scaleto{\mathrm{L}}{4pt}}(\mathbf{z}_{j^*} \mid \mathbf{x},\mathbf{d}_j)\log \frac{q_{\scaleto{\mathrm{L}}{4pt}}(\mathbf{z}_{j^*} \mid \mathbf{x},\mathbf{d}_j)}{p_{\scaleto{\mathrm{R}}{4pt}}(\mathbf{d}_j \mid \mathbf{z}_{j^*})}.
\end{equation}
This loss is exclusively used to optimize the retriever's parameters, without affecting the LM.

\textbf{Generation loss.}
Our technique for training LM utilizes the attention scores of the LM to train the LM itself, which is inspired by previous works of open-domain QA that train a retriever by learning to approximate the attention scores of the reader~\citep{distill-IzacardG21,atlas}. The cross-attention scores between input and output can be used as a proxy of the usefulness of each input to the justification. We firstly average decoder cross-attention scores over all attention heads, layers, and tokens for each retrieved document $\mathbf{d}_j$, resulting an averaged attention score $a(\mathbf{x}\oplus\mathbf{d}_j)$, where $\oplus$ denotes concatenation. Then the score that indicates the usefulness of $\mathbf{d}_j$ is obtained by applying the softmax operator $p(\mathbf{d}_j)=\frac{\text{exp}(a(\mathbf{x}\oplus\mathbf{d}_j))}{\sum_{k=1}^{N}\text{exp}(a(\mathbf{x}\oplus\mathbf{d}_k))}$ following~\citet{atlas}.
Similarly, the score for each chunk $\mathbf{z}_{i}$ is $p(\mathbf{z}_i)$, while the score of the most similar chunk $\mathbf{z}_{j^*}$ to $\mathbf{d}_j$ is $p'(\mathbf{z}_{j^*})=\frac{\text{exp}(p(\mathbf{z}_{j^*}))}{\sum_{k=1}^{N}\text{exp}(p(\mathbf{z}_{k^*}))}$. The objective is to encourage the score of $\mathbf{d}_j$ to approximate the score of its most similar chunk $\mathbf{z}_{j^*}$. We then minimize the KL-divergence between distributions of these two scores:  
\begin{equation}   
\mathcal{L}^{\mathrm{lm}}_{c}=\sum_{j=1}^{N} p'(\mathbf{z}_{j^*})\log \frac{p'(\mathbf{z}_{j^*})}{p(\mathbf{d}_j)}.
\end{equation}

\begin{table*}[t!]
\small
\begin{adjustbox}{width={\linewidth},keepaspectratio}%
    \begin{tabular}{lllccccc}
    \toprule[1.0pt]
      \textbf{Method} &\#Para.  & Test & ROUGE-1  & ROUGE-2   & ROUGE-L    & SummaCC   & MAUVE \\
    \midrule[1.0pt]
    \multirowcell{2}[0pt][l]{$\textbf{ExplainMT}_{\text{Full-dataset}}$\\\cite{atanasova-etal-2020-generating-fact}}  & \multirow{2}{*}{132M}  & F.C. Article & $35.01_{(-)}$ & $22.13_{(-)}$ & $21.25_{(-)}$ & $22.70_{(-)}$   & $5.59_{(-)}$ \\
      & &  Retr. Docs & $19.33_{(-)}$ & $9.55_{(-)}$ & $17.59_{(-)}$ & $9.34_{(-)}$   & $5.27_{(-)}$ \\
    \midrule[0.5pt]
     \multirowcell{2}[0pt][l]{$\textbf{ExplainerFC}_{\text{Full-dataset}}$\\\cite{kotonya-toni-2020-explainable-automated}}  & \multirow{2}{*}{340M} & F.C. Article & $62.10_{(-)}$ & $38.03_{(-)}$ & $54.25_{(-)}$ & $50.67_{(-)}$  & $14.63_{(-)}$ \\
     & & Retr. Docs & $47.16_{(-)}$ & $24.88_{(-)}$ & $44.13_{(-)}$ & $35.82_{(-)}$  & $10.07_{(-)}$ \\
      
     \midrule[0.5pt]
     \multirowcell{2}[0pt][l]{$\textbf{Atlas}_{\text{Few-shot}}$\\\cite{atlas}} & \multirow{2}{*}{$\sim$3B}  & F.C. Article & $40.93_{(0.97)}$ & $26.71_{(1.15)}$  &$33.98_{(1.01)}$   & $29.72_{(1.22)}$  &$28.25_{(2.46)}$ \\
      &  & Retr. Docs & $28.14_{(0.87)}$ & $13.91_{(1.31)}$  &$21.87_{(1.12)}$   & $12.64_{(0.87)}$   &$25.37_{(0.69)}$ \\
    \bottomrule[1.0pt]
    \end{tabular}
    \end{adjustbox}
    \caption{Results of justification generation methods trained on Fact-check Article (F.C. Article) and tested on Fact-check Article / Retrieved Documents (Retr. Docs). Para.: Parameters. Standard deviation is in (.).}
\label{tab:fc exp}
\end{table*}

\begin{table*}[t!]

    \begin{subtable}{\textwidth}
    \small
    \centering
    \begin{adjustbox}{width={\linewidth},keepaspectratio}%
        \begin{tabular}{llcccccc}
        \toprule[1.0pt]
         & \#Parameters & ROUGE-1 & ROUGE-2   & ROUGE-L   & SummaCC    & MAUVE  \\ 
        \midrule[0.5pt]
        \rowcolor{lightgray}
        \multicolumn{7}{l}{\textbf{Lead-4}~\cite{lead-4}}\\
        \specialrule{0em}{1pt}{1pt}
        & - & $22.72_{(-)}$ & $5.72_{(-)}$  & $14.11_{(-)}$   & $2.26_{(-)}$   & $7.95_{(-)}$ \\
        \rowcolor{lightgray}
        \multicolumn{7}{l}{\textbf{Retriever + ICL-enabled LMs}}\\
        \specialrule{0em}{1pt}{1pt}
        BM25~\cite{Okapi-BM25}\\
        \hspace{3mm} + Flan-T5~\cite{flan-t5}
                              & 11B & $27.99_{(2.39)}$ & $14.14_{(1.06)}$  &$20.74_{(1.66)}$   & $14.55_{(0.90)}$   & $12.42_{(1.22)}$\\
        \specialrule{0em}{1pt}{1pt}
        \hspace{3mm} + Llama2~\cite{touvron2023LLAMA}
                            &70B & $31.45_{(0.51)}$ & $12.36_{(0.25)}$  &$20.72_{(0.22)}$   & $13.05_{(0.38)}$   & $7.88_{(0.15)}$\\ 
        \specialrule{0em}{1pt}{1pt}
        \hspace{3mm} + GPT-4~\cite{gpt4}
                              & Unkown & $\mathbf{39.72}_{(1.97)}$ & $17.12_{(1.97)}$  &$26.18_{(2.26)}$   & $\mathbf{24.98}_{(2.49)}$   & $14.73_{(2.86)}$\\ 
        \midrule[0.5pt]
        Contriever~\cite{contriever}\\
        \hspace{3mm} + Flan-T5
                              & $\sim$11B & $23.75_{(1.91)}$ & $11.34_{(1.17)}$  &$18.11_{(1.48)}$   & $9.93_{(0.29)}$  & $12.07_{(0.90)}$\\
        \specialrule{0em}{1pt}{1pt}
        \hspace{3mm} + Llama2
                            &$\sim$70B & $31.28_{(0.51)}$ & $11.52_{(0.82)}$  &$20.42_{(0.70)}$   & $11.06_{(0.14)}$  & $7.91_{(0.09)}$\\ 
        \specialrule{0em}{1pt}{1pt}
        \hspace{3mm} + GPT-4
                              & Unkown & $\underline{36.83}_{(1.37)}$  & $14.10_{(1.66)}$  &$23.36_{(1.75)}$   & $\underline{20.07}_{(2.37)}$  & $9.85_{(0.96)}$\\ 
        \midrule[0.5pt]
        \rowcolor{lightgray}
        \multicolumn{7}{l}{\textbf{Atlas}~\cite{atlas}}\\
        \specialrule{0em}{1pt}{1pt}
        No joint training  & 3B & $31.42_{(1.61)}$ & $16.53_{(0.86)}$  &$24.67_{(1.00)}$   & $13.55_{(0.54)}$  & $25.19_{(4.37)}$\\ 
        \specialrule{0em}{1pt}{1pt}
        Joint training
                            & $\sim$3B  & $31.91_{(1.78)}$ & $17.81_{(1.19)}$  &$25.60_{(1.16)}$   & $13.81_{(1.11)}$  & $25.51_{(2.08)}$\\ 
        \midrule[0.5pt]
        \rowcolor{lightgray}
        \multicolumn{7}{l}{\textbf{JustiLM (Ours)}} \\
        \specialrule{0em}{1pt}{1pt}
         \hspace{3mm} $\mathcal{L}^{\mathrm{ret}}_{g}$ + $\mathcal{L}^{\mathrm{lm}}_{g}$ & $\sim$3B & $33.48_{(1.33)}$ & $18.59_{(0.79)}$  &$27.12_{(0.81)}$   & $15.04_{(1.27)}$   & $20.29_{(2.00)}$ \\
        \specialrule{0em}{1pt}{1pt}
        \hspace{3mm} $\mathcal{L}^{\mathrm{ret}}_{g}$ + $\mathcal{L}^{\mathrm{lm}}_{c}$ & $\sim$3B & ${36.70}_{(0.77)}$  & $\textbf{19.23}_{(0.84)}$  &$\textbf{28.39}_{(0.75)}$   & $14.80_{(0.45)}$   & $32.99_{(3.33)}$ \\  
        \specialrule{0em}{1pt}{1pt}
        \hspace{3mm} $\mathcal{L}^{\mathrm{ret}}_{c}$ + $\mathcal{L}^{\mathrm{lm}}_{g}$ & $\sim$3B & $36.51_{(1.01)}$  & $18.67_{(1.00)}$  &${27.94}_{(0.96)}$   & $14.77_{(0.19)}$  & $\textbf{37.08}_{(1.53)}$ \\
        \specialrule{0em}{1pt}{1pt}
        \hspace{3mm} $\mathcal{L}^{\mathrm{ret}}_{c}$ + $\mathcal{L}^{\mathrm{lm}}_{c}$  & $\sim$3B & $36.30_{(0.91)}$  & $\underline{18.68}_{(0.96)}$  &$\underline{27.97}_{(0.99)}$   & $14.69_{(0.48)}$  & $\underline{35.30}_{(1.09)}$ \\  
        \bottomrule[1.0pt]
        \end{tabular}
    \end{adjustbox}
    \subcaption{On the original test set with 987 claims indicated in Table~\ref{tab:dataset statistics}.}
    \label{tab:experiments}
    \end{subtable}
 
  \vspace{0.2cm}

  \begin{subtable}{\textwidth}
    \small
    \centering
    \begin{adjustbox}{width={\linewidth},keepaspectratio}%
        \begin{tabular}{llccccc}
        \toprule[1.0pt]
         & \#Parameters & ROUGE-1 & ROUGE-2   & ROUGE-L   & SummaCC  & MAUVE  \\ 
        \midrule[0.5pt]
        \rowcolor{lightgray}
        \multicolumn{7}{l}{\textbf{Lead-4}~\cite{lead-4}}\\
        \specialrule{0em}{1pt}{1pt}
        & - & $21.87_{(-)}$ & $3.95_{(-)}$  &$12.61_{(-)}$   & $1.70_{(-)}$  & $6.62_{(-)}$\\
        \rowcolor{lightgray}
        \multicolumn{7}{l}{\textbf{Retriever + ICL-enabled LMs}}\\
        \specialrule{0em}{1pt}{1pt}
        BM25~\cite{Okapi-BM25}\\
        \hspace{3mm} + Flan-T5~\cite{flan-t5}
                              & 11B & $22.86_{(2.27)}$  & $7.63_{(0.70)}$  &$14.74_{(1.52)}$   & $10.94_{(2.37)}$ & $7.00_{(0.17)}$\\
        \specialrule{0em}{1pt}{1pt}
        \hspace{3mm} + Llama2~\cite{touvron2023LLAMA}
                            &70B & $31.01_{(0.29)}$  & $9.64_{(0.32)}$  &$18.73_{(0.17)}$   & $11.49_{(0.69)}$   & $6.99_{(0.60)}$\\ 
        \specialrule{0em}{1pt}{1pt}
        \hspace{3mm} + GPT-4~\cite{gpt4}
                              & Unkown & $\mathbf{38.28}_{(1.44)}$ & $13.74_{(1.75)}$  &$23.36_{(2.20)}$   & $\mathbf{25.10}_{(2.29)}$  & $7.47_{(1.30)}$\\ 
        \specialrule{0em}{1pt}{1pt}
        \midrule[0.5pt]
        Contriever~\cite{contriever}\\
        \hspace{3mm} + Flan-T5
                              & $\sim$11B & $20.44_{(1.27)}$ & $7.93_{(0.48)}$  &$14.45_{(0.85)}$   & $10.18_{(2.03)}$   & $8.24_{(0.48)}$\\
        \specialrule{0em}{1pt}{1pt}
        \hspace{3mm} + Llama2
                            &$\sim$70B & $31.01_{(0.84)}$ & $9.81_{(0.73)}$  &$19.07_{(0.63)}$   & $10.75_{(0.52)}$   & $6.62_{(0.54)}$\\ 
        \specialrule{0em}{1pt}{1pt}
        \hspace{3mm} + GPT-4
                              & Unkown & $\underline{35.93}_{(1.09)}$ & $12.07_{(1.51)}$  &$21.46_{(1.79)}$   & $\underline{21.79}_{(2.22)}$  & $6.25_{(0.37)}$\\ 
        \midrule[0.5pt]
        \rowcolor{lightgray}
        \multicolumn{7}{l}{\textbf{Atlas}~\cite{atlas}}\\
        \specialrule{0em}{1pt}{1pt}
        No joint training
                              & 3B & $29.76_{(0.98)}$ & $13.40_{(0.34)}$  &$22.16_{(0.32)}$   & $10.78_{(0.55)}$  & $12.56_{(1.56)}$\\ 
        \specialrule{0em}{1pt}{1pt}
        Joint training
                                & $\sim$3B & $30.78_{(1.95)}$ & $15.75_{(1.72)}$  &$23.84_{(1.48)}$   & $12.20_{(0.45)}$   & $14.09_{(2.34)}$\\ 
        \midrule[0.5pt]
        \rowcolor{lightgray}
        \multicolumn{7}{l}{\textbf{JustiLM (Ours)}} \\
        \specialrule{0em}{1pt}{1pt}    
         \hspace{3mm} $\mathcal{L}^{\mathrm{ret}}_{g}$ + $\mathcal{L}^{\mathrm{lm}}_{g}$  & $\sim$3B & $32.76_{(0.89)}$ & $17.40_{(0.65)}$  &$26.61_{(0.61)}$   & $14.75_{(1.45)}$   & $10.57_{(0.94)}$ \\
        \specialrule{0em}{1pt}{1pt}
        \hspace{3mm} $\mathcal{L}^{\mathrm{ret}}_{g}$ + $\mathcal{L}^{\mathrm{lm}}_{c}$ & $\sim$3B & $35.55_{(0.31)}$  & $\textbf{17.84}_{(0.48)}$  &$\textbf{27.30}_{(0.21)}$   & $14.11_{(1.40)}$   & $16.78_{(4.64)}$ \\  
        \specialrule{0em}{1pt}{1pt}
        \hspace{3mm} $\mathcal{L}^{\mathrm{ret}}_{c}$ + $\mathcal{L}^{\mathrm{lm}}_{g}$ & $\sim$3B & $35.51_{(0.51)}$  & $17.21_{(0.70)}$  &$26.53_{(0.06)}$   & $14.30_{(0.40)}$  & $\textbf{20.02}_{(7.39)}$ \\
        \specialrule{0em}{1pt}{1pt}
        \hspace{3mm} $\mathcal{L}^{\mathrm{ret}}_{c}$ + $\mathcal{L}^{\mathrm{lm}}_{c}$  & $\sim$3B & $35.48_{(0.59)}$  & $\underline{17.52}_{(0.86)}$  &$\underline{26.92}_{(0.57)}$   & $13.99_{(0.49)}$  & $\underline{19.17}_{(7.04)}$ \\  
        \bottomrule[1.0pt]
        \end{tabular}
    \end{adjustbox}
    \subcaption{On the new test set with 348 claims published later than the claims from the WatClaimCheck dataset used for training.}
    \label{tab:new test}
    \end{subtable}

  \caption{Few-shot justification generation results on test set (a) and new test set (b). Standard deviation is in (.).}
  \label{tab:main results}
\end{table*}

\section{Experiments and Results}

\subsection{Evaluation Metrics}
To assess the consistency of generated justifications with ground truth, we employ a spectrum of metrics to make our evaluation balance between factual accuracy and style diversity of verbal expressions: \textbf{ROUGE}~\cite{lin-2004-rouge} counts the number of overlapping units (e.g., n-gram and word sequences) between output justifications and ground truths. 
\textbf{MAUVE}~\cite{mauve} measures the divergence between output justifications and the ground truths, which could reflect whether the output is fluent and coherent to the ground~\citep{conf-aaai-Xie0MM23,krishna-etal-2022-rankgen,Enabling-Citations,xu-etal-2023-best}. 
\textbf{SummaCC} expands the SummaC~\cite{laban-etal-2022-summac} to evaluate the \underline{c}overage and factual \underline{c}onsistency through checking entailment between the output justifications and ground truth. It sums the aggregating NLI scores over the pairs of the entire output justification and each sentence in the ground truth for coverage~\cite{scialom-etal-2021-questeval,Enabling-Citations}, and reversely, the pairs of the entire ground truth justification and each sentence in the output justification for consistency~\cite{laban-etal-2022-summac}.

\subsection{Fallacy of Fact-Check Summarization}
We investigate how previous approach based on fact-check article summarization~\citep{kotonya-toni-2020-explainable-automated,atanasova-etal-2020-generating-fact} fails to generalize to the realistic setting given retrieved evidence rather than fact-check articles as input. 

\textbf{Experimental Setup.} 1) Full training: we include two existing models, ExplainMT~\cite{atanasova-etal-2020-generating-fact} and ExplainerFC~\cite{kotonya-toni-2020-explainable-automated}. ExplainMT is an extractive model while ExplainerFC is extractive-abstractive. We partition the training set of ExClaim into 5,000 instances for training and 964 for validation. We train the two models to summarize fact-check articles, and test them by inputting fact-check articles versus evidence documents retrieved with BM25~\cite{Okapi-BM25}. 2) Few-shot training: we train the RAG model Atlas~\cite{atlas} under few shots with fact-check articles as input and test it using fact-check articles versus documents retrieved by its pre-trained retriever Contriever. In this setting, Contriever will be fixed during fine-tuning since the LM's input is fact-check articles. We use randomly sampled 30 shots from the training split, and report the results averaged over 3 trials based on different seeds.

\textbf{Results.} As shown in Table \ref{tab:fc exp}, for both settings, we observe that using retrieved documents as input dramatically declines the performance compared to inputting fact-check articles. This suggests that the fact-check article summarization approach struggles to generalize to the retrieved documents, especially in few-shot setting, indicating the impracticality of previous approaches and the importance of the more realistic framework outlined in $\S$\ref{Task Formulation}. That is, models need to generate justifications based on retrieved evidence instead of fact-check articles which are not available for new claims during inference.

\subsection{Few-shot Justification Generation}

\subsubsection{Baselines}
1) \textbf{Lead-4}~\cite{lead-4} selects as justification the first sentence from each document among the top-4 documents retrieved by BM25. 
2) \textbf{Retriever + ICL-enabled LMs:} We use BM25 as the sparse retriever and Contriever~\cite{contriever} as the dense retriever, and choose Flan-T5 (11B)~\cite{flan-t5},  Llama2 (70B)~\cite{touvron2023LLAMA} and GPT-4~\cite{gpt4} as the ICL-enabled LMs.
We prompt the model to generate justifications by concatenating few-shot training instances along with a test instance. 
3) \textbf{Atlas}~\cite{atlas} is the SoTA RAG model with strong few-shot ability, which consists of a trainable dense retriever Contriever and a LM-adapted variant of T5~\cite{lester-etal-2021-power} with Fusion-in-Decoder~\cite{izacard-grave-2021-leveraging} modified to increase the number of retrieved documents. We also include a non-joint training setting by replacing the retriever with BM25.

\subsubsection{Experimental Setup}
For our method JustiLM, we randomly sample 30 instances from the training set for fine-tuning. We use the Atlas~\cite{atlas} with its released pre-trained checkpoint\footnote{\url{https://github.com/facebookresearch/atlas}} of 3B parameters as our backbone model. Following the Atlas paper, we retrieve top-20 documents for each instance. We set training steps as 100, batch size as 8, and learning rate as \num{4e-5} with linear decay and 5 warmup steps for both the LM and the retriever. 

For the distillation techniques to train the LM, we begin by fine-tuning the LM to take fact-check articles as auxiliary input and generate justification, which provides a warmup for LM. 
For BM25 + ICL-enabled LMs, we use the Pyserini\footnote{\url{https://github.com/castorini/pyserini}} toolkit to build BM25 model. For Flan-T5, We use the code and pre-trained checkpoints from Hugging Face Transformers\footnote{\url{https://huggingface.co/google/flan-t5-xxl}}. We use the original code and pre-trained checkpoints of Llama2\footnote{\url{https://github.com/facebookresearch/LLAMA}}. We use the API service of GPT-4 from OpenAI\footnote{\url{https://openai.com/gpt-4}}. 
Given different length constraints of these LMs, we intend to maximize the utilization of their specific input capabilities. We adjust the number of the shots and/or the number of retrieved documents to maximally utilize their input context windows. We prioritize to ensure that these models have access to as many of the top-20 retrieved documents as possible because effective generation requires an adequate amount of information, with the secondary goal to maximize the number of few-shot examples used. Specifically, we set 1-shot ICL with top-10 documents for Flan-T5, 2-shot ICL with top-20 documents for Llama2 and 3-shot with top-20 documents for GPT-4. 

For fair and robust comparison, we perform experiments three times, with training instances sampled using different random seeds. We report the mean and standard deviation of each metric over the three runs in all experiments. The seeds and training instances are kept the same across different models. All the experiments use a server with 8 NVIDIA Tesla-V100 32GB GPUs.

\begin{table*}[t]
\small
\centering
\begin{adjustbox}{width={0.9\linewidth},keepaspectratio}%
    \begin{tabular}{c|c|ccccc}
    \toprule[1.0pt]
     Component & Loss  & ROUGE-1  & ROUGE-2   & ROUGE-L    & SummaCC   & MAUVE \\
    \midrule[1.0pt]
    \multirowcell{2}[0pt][l]{Retriever} & $\mathcal{L}^{\mathrm{ret}}_{g}$  &  $32.13_{(0.99)}$ & $16.45_{(0.39)}$  &$25.15_{(0.59)}$   & $14.53_{(0.23)}$  & $26.53_{(3.38)}$\\  
    \cmidrule{2-7}
    & $\mathcal{L}^{\mathrm{ret}}_{c}$  & $31.29_{(1.53)}$ & $17.26_{(0.94)}$  &$25.19_{(1.15)}$   & $13.77_{(1.41)}$   & $19.17_{(1.57)}$\\ 
    \midrule[1.0pt]
    \multirowcell{2}[0pt][l]{LM} & $\mathcal{L}^{\mathrm{lm}}_{g}$  & $36.30_{(1.80)}$ & ${19.23}_{(1.05)}$  &${28.52}_{(1.04)}$   & $14.92_{(0.73)}$  & ${27.09}_{(7.20)}$\\ 
    \cmidrule{2-7}
    & $\mathcal{L}^{\mathrm{lm}}_{c}$  & ${37.03}_{(0.80)}$ & ${18.89}_{(0.90)}$  &${28.29}_{(0.79)}$   & $14.56_{(0.69)}$   & ${34.16}_{(3.73)}$ \\   
    \bottomrule[1.0pt]
    \end{tabular}
    \end{adjustbox}
    \caption{Results of ablations on different distillation techniques. (.) encloses standard deviation.}
\label{tab:ablation}
\end{table*}

\begin{table*}[t!]
\small
\centering
\begin{adjustbox}{width={0.9\linewidth},keepaspectratio}%
    \begin{tabular}{l|l|cccccc}
    \toprule[1.0pt]
     \textbf{Method} & macro-F1  & ROUGE-1  & ROUGE-2   & ROUGE-L    & SummaCC   & MAUVE \\
      \midrule[1.0pt]
    \textbf{Majority} & $23.34_{(-)}$  & - & -  & -   & -   & - \\
    \midrule[1.0pt]
    \textbf{Atlas-CLS} & $25.81_{(0.46)}$ & -  & -  & -   & -   & - \\
    \midrule[0.5pt]
        \textbf{JustiLM}-${\mathcal{L}^{\mathrm{lm}}_{g}}$ & ${44.00}_{(1.51)}$ & ${32.52}_{(1.39)}$ & ${18.20}_{(0.61)}$  &${26.34}_{(0.88)}$    & ${14.76}_{(1.17)}$   & ${18.68}_{(1.80)}$ \\
     \specialrule{0em}{1pt}{1pt}
        \textbf{JustiLM}-$\mathcal{L}^{\mathrm{lm}}_{c}$ & ${41.33}_{(4.49)}$  & ${35.87}_{(1.02)}$  & ${19.52}_{(0.68)}$ &${28.22}_{(0.86)}$   & ${15.02}_{(0.95)}$   & ${32.98}_{(2.96)}$ \\  
    \bottomrule[1.0pt]
    \end{tabular}
    \end{adjustbox}
    \caption{Results of joint veracity prediction and justification generation. (.) encloses standard deviation.}
\label{tab:mt}
\end{table*}

\subsubsection{Main Results} \label{sec:main results}
The results of few-shot justification generation methods are reported in Table \ref{tab:experiments}. 
Lead-4 that directly presents the retrieved documents as justification does not yield satisfactory results, due to simple evidence stacking without generating a clear explanation of the rationale. 

Both Flan-T5 and Llama2 outperform Lead-4, demonstrating the LM's ability to generate justifications based on retrieved evidence. Flan-T5 performs comparably with Llama2 in ROUGE and SummaCC scores and better in MAUVE, despite much fewer parameters. The reasons are likely two-fold: 1) Flan-T5's instruction fine-tuning on 1.8K tasks, which effectively enhances the pre-trained language models~\cite{sanh2022multitask,flan-t5}; 2) its fine-tuning on Chain-of-Thought (CoT) data~\cite{cot}, aligning with the common presentation of ground-truth justifications that provide rationales to conclude the veracity, as exemplified in Table \ref{tab:intro example}.

Incorporating ICL-enabled LMs with the dense retriever Contriever does not exhibit improvement over using the sparse retriever BM25. Dense retrievers that trained on extensive in-domain training datasets like MS-MARCO~\cite{MS-MARCO}, are often surpassed by sparse retrievers when applied to new domains without large annotated datasets~\citep{thakur2021beir,contriever}. While Contriever is a strong unsupervised retriever for bridging this gap, BM25 still remains competitive~\citep{contriever}. 

When training only the LM of Atlas, it demonstrates superior overall performance compared to Flan-T5 and Llama2, despite its much fewer parameters. This finding indicates that merely relying on the implicit knowledge of LMs without parameter updates is insufficient when the size of LM is not large enough. Joint training of the retriever and LM leads to further performance gains, implying its benefits in the few-shot setting.

Compared to Atlas, JustiLM makes improvements in different metrics, indicating that utilizing fact-check article as auxiliary training signals enhances justification quality. With our proposed distillation techniques, JustiLM considerably improves all ROUGE scores. Compared to Atlas, the combination of article-level distillation on retriever and chunk-level distillation on LM increases ROUGE-1, ROUGE-2, and ROUGE-L scores by 15.0\%, 7.97\% and 10.9\%, respectively, suggesting that JustiLM can generate justifications which are more similar to those written by fact-checkers. Furthermore, 3 out of 4 combinations of distillation techniques outperform Atlas in MAUVE scores, with the highest gain being 45\%. This suggests that JustiLM's justifications are more fluent and coherent with ground truths. It can be attributed to our distillation method allowing the model to learn from fact-check articles that are much more informative and detailed than the explanatory justifications. Lastly, JustiLM effectively enhances the SummaCC score, indicating the improvements on the factual consistency of generated justifications. 

GPT-4 demonstrates exceptionally strong ability in providing factually consistent responses and outperforms other ICL-enabled methods Flan-T5 and Llama2 across all metrics. 
In comparison, JustiLM falls relatively below GPT-4 in ROUGE-1 and SummaCC, but outperforms GPT-4 in ROUGE-2/L and MAUVE. This highlights its effectiveness, especially considering its small model size and independence from intensive compute and storage resources required by very large models. Also, its ease of fine-tuning with more and new training data provides significant flexibility in addressing the ever-changing landscape of misinformation.

\subsubsection{Generalization on New Claims} \label{sec: new test}
To address the concern of pre-trained LMs having potentially seen the evaluation data during their pre-training, we investigate how different methods perform on a new test set with new emerging claims made after their training. Since the WatClaimCheck dataset exclusively encompasses claims prior to July 2021~\cite{khan-etal-2022-watclaimcheck} and the newest pre-training data of Llama2 are cut off by September 2022, we gather a new set of claims made between October 2022 and September 2023, yielding a new test set comprising 348 instances, each with their associated reference documents and justifications. Following the same steps detailed in \S \ref{sec:dataset}, the newly collected reference documents are added into the corpus for model retrieval. As shown in Table \ref{tab:new test}, all methods demonstrate performance drop on the new test set. Nonetheless, the findings obtained based on the original test set still hold true for the new test data. Additionally, compared to baseline methods, the relatively mild performance drop in JustiLM suggests stronger generalizability and robustness of our distillation techniques.

\subsubsection{Ablation on Distillation Techniques}
Table \ref{tab:ablation} reports the result of ablations on our distillation techniques. We observe that the distillation during LM training results in greater improvements compared to the retriever. This is expected, considering that the LM benefits from direct supervision from ground-truth justifications during training, while the retriever relies on the weak supervision from LM and the distillation of fact-check articles. Additionally, the LM has a larger number of parameters than the retriever, with 3 billion parameters for the LM compared to 110 million parameters for the retriever. As a result, the LM tends to capture more knowledge from fact-check article during the distillation process, leading to substantial improvements in performance.

\begin{table*}[t!]
    \small
    \begin{tabularx}{\linewidth}{lX}
    \toprule
    \textbf{Claim} & Says Wisconsin women facing pay discrimination can't "do something about it" under bill passed by Republicans. \\
    \specialrule{0em}{1pt}{1pt}
    \textbf{Ground Truth} &  Falk said women facing pay discrimination can't "do something about it" under a bill passed by Wisconsin Republicans. But of the four legal options available under the current law, only one would be eliminated by the bill. We rate Falk's statement False. \\
    \specialrule{0em}{1pt}{1pt}
    \textbf{Atlas} &  Falk said that under a Republican-backed bill, women facing pay discrimination can't "do something about it." Falk said that under the GOP bill, women facing pay discrimination can't "do something about it." But the bill doesn't change the law. It simply makes it harder for women to file lawsuits. Falk said that under the GOP bill, women facing pay discrimination can't "do something about it." \\
    \specialrule{0em}{1pt}{1pt}
    \textbf{GPT-4} &  A statement by Kathleen Falk, a former Dane County executive, claims that women in Wisconsin would no longer have "a right to do something" about pay discrimination on the job if a bill is signed by Governor Walker. The bill in question would take away the right of women who suffer pay discrimination to sue in state court. However, there are still other avenues for employment discrimination victims, including filing a complaint with the state Equal Rights Division, filing a complaint with the federal Equal Employment Opportunity Commission, and filing a lawsuit in federal court. Therefore, while the bill may limit one avenue for action, it does not completely prevent women from taking action against pay discrimination. \\
    \specialrule{0em}{1pt}{1pt}
    \textbf{JustiLM} & Falk said that women facing pay discrimination can't "do something about it" under a bill passed by Republicans. The measure would make it harder on victims of employment discrimination because it is easier to sue in state court than in federal court. But eliminating the state lawsuit option wouldn't affect the three other avenues of legal recourse available to workplace discrimination victims. And Falk's claim was women facing pay discrimination would have no options at all.  \\
    \bottomrule
    \end{tabularx}
    \caption{An Example of generated justifications by different methods compared to the ground-truth justification.}
    \label{tab:case}
\end{table*}

\subsection{Joint Veracity-Justification Performance}
In this section, we demonstrate that JustiLM can be easily extended for joint veracity prediction and justification generation. We follow ~\citet{khan-etal-2022-watclaimcheck} to map the original veracity labels assigned by fact-checking websites, resulting in 388, 532 and 67 instances for the false, mixture and true classes in the test split, respectively. Such class imbalance is consistent with the report by~\citet{khan-etal-2022-watclaimcheck}. To mitigate the impact of imbalanced class distribution, we balance the 30 training shots across the three classes by randomly sampling 10 instances per class from the training set.

We make the LM generate the justification and veracity label at the same time. For veracity label prediction, let $y_{\mathrm{cls},i}$ be a veracity label, and its predicted score assigned by the LM conditioned on the claim and the retrieved documents is defined as $\beta(\mathbf{x},y_{\mathrm{cls},i},D_N)=\frac{1}{|y_{\mathrm{cls},i}|} \log p_{\scaleto{\mathrm{L}}{4pt}}(y_{\mathrm{cls},i} \mid \mathbf{x},D_N)$ following~\citet{liu2022few}. In this way, we rank all classes by the predicted scores and select the top-ranked class. During training, we calculate the probability of prediction by applying $\mathrm{softmax}$ function on the predicted scores, and use cross-entropy as the loss function.

Table \ref{tab:mt} presents the result. The Atlas-CLS, which directly predicts veracity label with Atlas, shows a limited improvement in macro-F1 score compared to the Majority method. This suggests that predicting the veracity of real-world claims remains challenging for this original RAG model in a few-shot setting. When performing joint veracity prediction and justification generation with the LM training, a substantial boost in verdict prediction is observed for our method. Specifically, we achieve absolute improvements of 18.19 and 15.52 in macro-F1 using article-level and chunk-level techniques, respectively. This indicates that justification generation can help veracity prediction by consolidating evidence from retrieved documents. We also find that jointly training JustiLM with the veracity prediction task does not improve the performance of justification generation, which is consistent with the findings by \citet{atanasova-etal-2020-generating-fact}. We conjecture that it remains challenging for the model to boost both tasks simultaneously with few-shot training instances. Potential solutions could consider either leveraging a larger multi-task training dataset, such as T0~\cite{sanh2022multitask}, or using an independent veracity classifier that can be jointly trained with the retriever and the LM. However, both options necessitate adding data and computational resources. We will leave it for future studies.

\subsection{Case Study}
Table \ref{tab:case} presents example justifications generated by JustiLM, the strong ICL baseline GPT-4, and the few-shot RAG model Atlas. Atlas's generated justification catches that the GOP bill does not change the law, but fails to highlight the key point that women still have viable avenues to address pay discrimination. Both GPT-4 and JustiLM successfully refute the claim by providing that crucial point. 

More specifically, Atlas falls short in delivering convincing and comprehensive justification due to its tendency to provide incomplete and repetitive responses. In contrast, GPT-4, being the SoTA LLM, impresses with its ability to generate well-rounded justification, but appears to be lengthy and less focused. JustiLM, on the other hand, successfully highlights key points for fact-checking the claim with a precise and refined justification. Despite its relatively small model size, JustiLM may not always offer the same level of details as GPT-4, but it can produce concise and accurate justifications that closely resemble the ground truth, making JustiLM promising and valuable for users seeking quick and trustworthy fact-check explanations.

\section{Discussion}
There is no passage-/sentence-level annotation in the original long-form reference documents and fact-check articles, which are costly to obtain. We do not have ground truths for training and evaluating evidence retrieval model. Since these long documents bury specific evidence in them, directly using them for training will introduce a considerable amount of irrelevant text. While we mitigate this challenge by splitting each original reference document into disjoint 100-word chunks for retrieval, we believe that acquiring fine-grained evidence annotations will benefit the training and evaluation.

In our experimental setup, evidence retrieval is conducted under the assumption that the needed evidence for fact-checking a given claim exists in the retrieval corpus. However, in real-world searching scenario where gold evidence may be absent from the retrieval corpus, it is valuable to investigate how justification generation methods perform under this more challenging scenario by varying the ratio of gold reference documents in the retrieval corpus.

Additionally, while our experiments include the NLI-based metric SummaCC, providing automated evaluation on the factuality of generated justifications, we believe that a sound human evaluation involving professional fact-checkers. Such evaluation, currently not conducted, necessitates close collaboration with fact-checking organizations and needs particular networking and setup, such as the integration with their existing workflow and the provision of motives for them to participate in evaluation, which could be warranted as a separate study by itself and is part of our future plan.

As the SoTA LLM, GPT-4 shows strong ability in generating factually consistent and informative justifications, therefore, developing justification methods based on those powerful API-based LLMs is beneficial. However, these blackbox LLMs have strict constraints on accessing their specific internal information, which poses important open challenges for being interacted with deeply and providing supervision signals to retriever.

In this work, we address the justification generation task with a realistic approach, which generates justifications based on the retrieved evidence using an end-to-end retrieval-augmented language model. Furthermore, incorporating our distillation techniques with the RAG model Atlas, demonstrates a marked improvement in performance. This affirms that utilizing fact-check articles during training to provide supervision signals can strongly enhance justification generation.

\section{Conclusion and Future Work}
We propose a justification generation language model JustiLM for realistic fact-checking of real-word news claims, where justification generation is performed based on retrieved evidence from large textual corpus,
and introduce a new benchmark dataset ExClaim for this task. JustiLM leverages fact-check articles as auxiliary resources during training to distill article-level and chunk-level training signals to guide justification writing. Experimental results show JustiLM outperforms ICL-enabled Flan-T5 and Llama2, as well as the SoTA few-shot RAG model Atlas. JustiLM also demonstrates comparable and promising performance when compared to GPT-4.  

In the future, we will explore the adaptation of various LLM-based reasoning methods (e.g., CoT~\cite{cot}, ToT~\cite{yao2023tree}, and GoT~\cite{besta2023graph}) into JustiLM to enhance the reasoning ability for improving the task of justification generation, which aims to assist the LMs in providing better signals for guiding evidence retrieval and improving reasoning over retrieved evidence during justification generation.
We also plan to develop a human evaluation scheme involving fact-checking experts to provide a more comprehensive and efficient assessment on machine-generated justifications.

\section*{Acknowledgments}
We would like to thank the anonymous reviewers and Action Editor Fei Liu for their helpful suggestions. We are also grateful to Alessandro Moschitti for his valuable comment and discussion.

\bibliography{tacl2021,anthology,custom}
\bibliographystyle{acl_natbib}

\iftaclpubformat







  
\fi

\end{document}